%% file: paper.tex
\documentclass[letterpaper, 10 pt, conference]{ieeeconf}  
\IEEEoverridecommandlockouts
\usepackage{graphicx}
\usepackage{times} 
\usepackage{amsmath} 
\usepackage{amssymb}  
\usepackage{units}
\usepackage{verbatim}  
\usepackage{nicefrac}		
\usepackage{acronym}		
\usepackage{glossaries}		
\usepackage[usenames,dvipsnames]{xcolor}			
\usepackage{cite} 
\usepackage{url} 
\usepackage[multidot]{grffile} 
\usepackage{multirow}
\usepackage{booktabs}

\makeatletter
\let\NAT@parse\undefined
\makeatother
\usepackage[ colorlinks=true, citecolor=black, linkcolor=black, urlcolor=magenta,  bookmarks=true, bookmarksnumbered=true, pdftex, plainpages=false, pdfpagelabels]{hyperref} 

\input{Acronyms}

\title{\LARGE \bf
NOCaL: Calibration-Free Semi-Supervised Learning of Odometry and Camera Intrinsics
}

\author{Ryan Griffiths, Jack Naylor, Donald G. Dansereau
\thanks{The authors are with the Australian Centre for Field Robotics, School of Aerospace, Mechanical and Mechatronic Engineering, The University of Sydney.
{\tt\small r.griffiths, jack.naylor, donald.dansereau@sydney.edu.au}}%
}

\input{commands}
\begin{document}
\maketitle
\thispagestyle{empty}
\pagestyle{empty}

\begin{abstract}
There are a multitude of emerging imaging technologies that could benefit robotics. However the need for bespoke models, calibration and low-level processing represents a key barrier to their adoption. In this work we present NOCaL, Neural Odometry and Calibration using Light fields, a semi-supervised learning architecture capable of interpreting previously unseen cameras without calibration. NOCaL learns to estimate camera parameters, relative pose, and scene appearance. It employs a scene-rendering hypernetwork pre-trained on a large number of existing cameras and scenes, and adapts to previously unseen cameras using a small supervised training set to enforce metric scale. We demonstrate NOCaL on rendered and captured imagery using conventional cameras, demonstrating calibration-free odometry and novel view synthesis. This work represents a key step toward automating the interpretation of general camera geometries and emerging imaging technologies. Code and datasets are available at \href{https://roboticimaging.org/Projects/NOCaL/}{https://roboticimaging.org/Projects/NOCaL/}.

\end{abstract}

\input{sections/01_introduction}

\input{sections/02_related}
\input{sections/03_methods}

\input{sections/04_results}

\input{sections/06_conclusions}

%

{\small
	\bibliographystyle{IEEEtran}
	\bibliography{./references}
}
\end{document}

%% file: Acronyms.tex
\newacronym{ACFR}{ACFR}{Australian Centre for Field Robotics}
\newacronym{USyd}{USyd}{the University of Sydney}
\newacronym{AUV}{AUV}{autonomous underwater vehicle}
\newacronym{UAV}{UAV}{unmanned aerial vehicle}
\newacronym{SLAM}{SLAM}{simultaneous localisation and mapping}
\newacronym{SfM}{SfM}{structure-from-motion}
\newacronym{SNR}{SNR}{signal-to-noise ratio}
\newacronym{DFT}{DFT}{discrete Fourier transform}
\newacronym{FFT}{FFT}{fast Fourier transform}
\newacronym{SIFT}{SIFT}{scale invariant feature transform}
\newacronym{TP}{TP}{true positive}
\newacronym{FP}{FP}{false positive}
\newacronym{BFF}{BFF}{burst feature finder}
\newacronym{5D}{5D}{five-dimensional}
\newacronym{4D}{4D}{four-dimensional}
\newacronym{3D}{3D}{three-dimensional}
\newacronym{2D}{2D}{two-dimensional}
\newacronym{1D}{1D}{one-dimensional}
\newacronym{NIR}{NIR}{near-infrared}
\newacronym{ORB}{ORB}{oriented FAST and rotated BRIEF}
\newacronym{SURF}{SURF}{speeded up robust features}
\newacronym{DoG}{DoG}{difference of Gaussians}
\newacronym{LiFF}{LiFF}{light field features}
\newacronym{ROC}{ROC}{receiver operating characteristic curve}

%% file: commands.tex
\newcommand{\myquat}[1]{\bar{#1}}

\newcommand{\q}{\myquat{q}}

\newcommand{\Cq}[2]{\boldsymbol{C}(\q)}



%% file: sections/01_introduction.tex
\section{Introduction}
\label{sec:intro}


Vision is a critical sense in modern robotics. Enormous advancements have been made in recent years to better utilise cameras for almost every task a robotic platform might be asked to do. Looking to nature, it is clear that the best way to see the world depends on what needs to be seen. In the same way, the best camera is specific to the application and domain in which a robot is operating. Whilst novel cameras are being developed to address shortcomings in existing modalities, this raises a key problem in deploying these sensors quickly and on new platforms: calibration and low-level interpretation.

Calibrating and interpreting new imaging devices is skilled and time-consuming work. Emerging devices like event cameras and light field cameras have taken years and even decades to adapt in robotics. Solutions generally involve the use of bespoke models, calibration procedures, and low-level interpretation and sensor fusion algorithms. An assumption is generally made of static camera characteristics, with device changes due to vibration, temperature, replacement or upgrading requiring re-calibration. This makes both integrating new cameras and managing fleets of robots onerous and complex.

\begin{figure}[t]
	\centering  
	\includegraphics[width=\columnwidth]{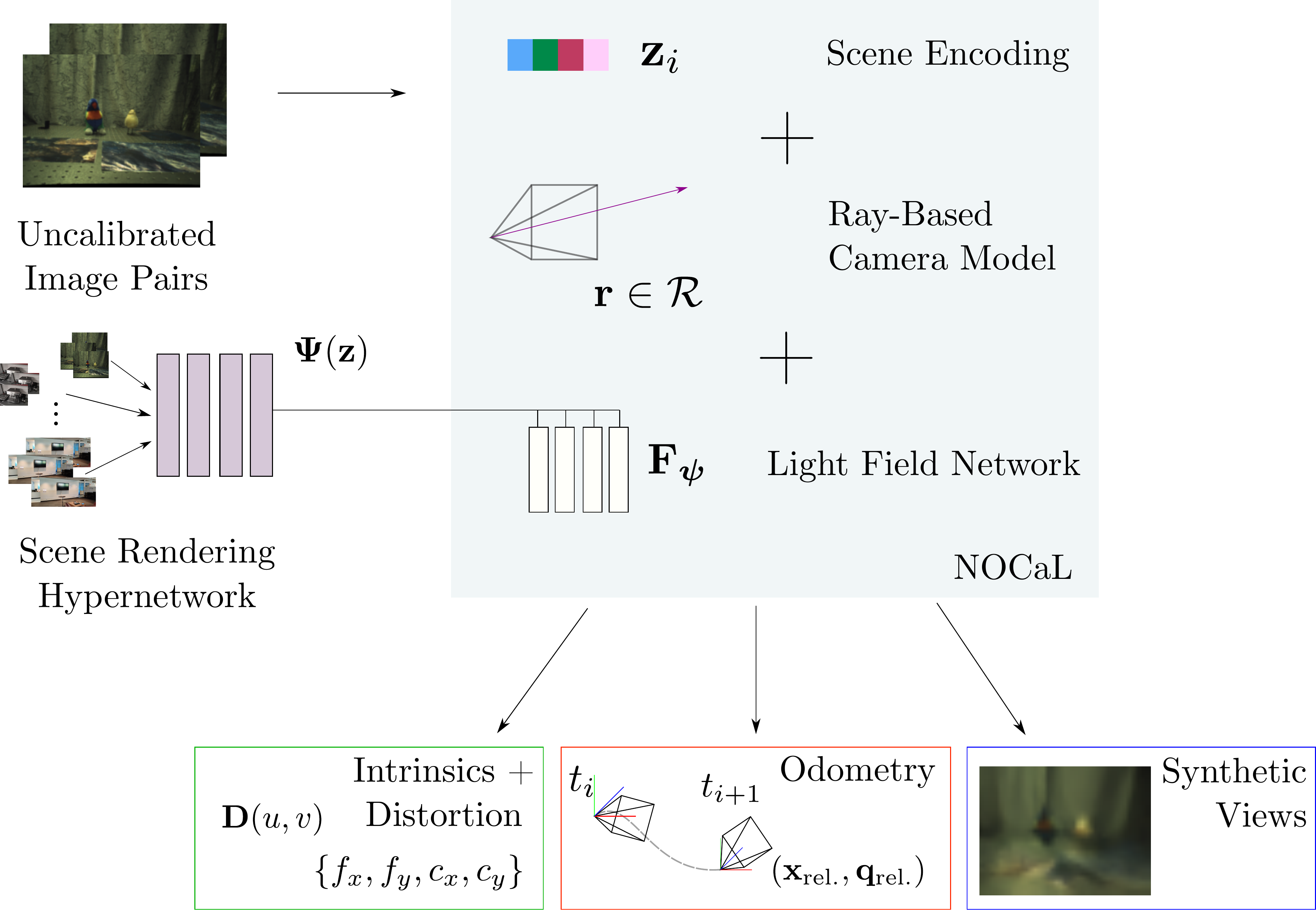}
	\caption{NOCaL learns odometry and novel view synthesis from previously unseen cameras by leveraging a rendering hypernetwork pre-trained on existing cameras. Uncalibrated input image pairs are encoded in a latent space $\mathbf{z}_i$ that drives the hypernetwork $\mathbf{\Psi(\mathbf{z})}$ to generate a light field rendering network $\mathbf{F}_{\boldsymbol{\psi}}$. A differentiable ray-based camera model drives the renderer, enabling estimation of camera parameters, 
    relative pose, and novel views. This work is a key step towards automatically interpreting more general camera geometries and emerging camera technologies.}\label{fig_abstract}
\end{figure}

In this work, we propose a framework to automatically interpret previously unseen cameras by jointly learning to estimate novel views, odometry, and camera parameters -- see Fig.~\ref{fig_abstract}. Our framework utilises recent advancements in neural rendering to provide self-supervision, leveraging the large amount of imagery available from a newly introduced uncalibrated camera. To benefit from the availability of unlabelled training data from existing cameras, we employ a hypernetwork that learns to construct light field renderers, so that the hypernetwork benefits from the large pool of existing data. Finally, to ground our odometry estimates in metric space, we employ a small labelled training set that we show is easy to collect where a complementary source of odometry is available.

To our knowledge, this is the first work to support automatic interpretation of previously unseen cameras without need for calibration. Through unsupervised learning of camera parameters and odometry, our approach benefits from the large amount of unlabelled data available from existing cameras as well as a newly introduced camera. By introducing a small labelled training set we constrain the solution to a metric space with known scale.

To demonstrate our approach, we employ cameras which are well described by a pinhole projection model with freeform ray-based distortion model. This allows use of a broad range of monocular cameras without calibration. In the future, we envision further relaxing this model to an entirely freeform ray-based model, opening the space to a broader range of cameras including stereo, multi-aperture and light field, and fisheye, that are all well described by ray-based geometry.


We validate our approach on both captured and rendered images of indoor scenes, using cameras with different focal lengths and distortion parameters. We demonstrate our system accurately estimating camera intrinsics, distortion models, and relative pose, i.e.~odometry. 

To position our work, we compare against a fully supervised approach and an unsupervised approach that requires calibration. Perhaps surprisingly, our semi-supervised but uncalibrated approach outperforms both in accuracy of odometry, demonstrating the strength of combining small labelled datasets with readily available unlabelled data. We also include an ablation study that establishes the importance of the distortion model when using cameras that deviate substantially from an ideal pinhole model, comparing against variants of our method that lack distortion model and that estimate no camera parameters at all.

We anticipate this work to find broad applicability where recalibration is difficult and camera parameters can change, either due to optical shifts or replacing of hardware. Deployed systems on planetary missions, in harsh environments, and in domestic applications like robotic vacuums for example are typically difficult to recalibrate. Changes to on-board calibration can occur due to vibration and thermal effects, and replacing or upgrading cameras can be an expensive proposition especially where new camera models and/or recalibration are required. Our framework requires no prior knowledge of hardware or camera parameters, allowing such robots to perform accurate camera pose estimation without need for recalibration or manual intervention. 

Limitations: Whilst our network is designed to work with a family of cameras, cameras not well-described by a pinhole projection and freeform distortion profile are unlikely to be well supported by our camera model. We anticipate generalising the camera model in future. Pose estimation requires substantial overlap between the input images, and so the approach also breaks down for fast motion / low-overlap input pairs.


%% file: sections/02_related.tex
\section{Related Work}
\label{sec:related}


State-of-the-art approaches to visual odometry jointly learn scene depth and odometry of monocular images using unsupervised learning~\cite{zhou2017unsupervised}. This uses a warp function that allows one image to be warped to another based on depth and estimated pose. While most studies improve odometry estimation using warp functions \cite{han2019deepvio, yin2018geonet, zhan2018unsupervised, godard2017unsupervised}, warping requires accurately calibrated images which are often not available on robotic platforms that are deployed in harsh environments. This also limits the number of cameras that can be deployed because novel cameras require long hours of calibration in controlled environment, with specialised targets and expertise. 

Digumarti et al.~\cite{digumarti2021unsupervised} demonstrated the performance of warping using a novel 4D warp function by extending it to a new family of cameras: sparse light-field cameras. In using a warp function, these studies are limited to scenes with simple well-explained phenomena, for example these methods cannot handle view dependent phenomena (reflection, refraction).

Recent studies on neural novel view synthesis~\cite{mildenhall2021nerf} have demonstrated state of the art results, with applications to many computer vision applications. While this has been adapted to robotics in \cite{adamkiewicz2022vision,Sucar:etal:ICCV2021,IchnowskiAvigal2021DexNeRF,yen2022nerfsupervision}, the fundamental limitation of such approaches exists in being only able to represent a specific scene, and within the spatial region captured by the input data. The computation required to ray-march is expensive, but provides a dense and continuous scene representation. Instant-NGP~\cite{muller2022instant} addresses this limitation to an extent by employing a multi-resolution hash encoding, which drastically speeds up training and inference, but leads to more sparse geometry.

Light Field Networks (LFNs)~\cite{sitzmann2021lfns} perform a single query of the network unlike prior works, enabling a reduction in training and inference time by several orders of magnitude compared to NeRF~\cite{mildenhall2021nerf}. Recent studies \cite{attal2022learning, li2022neulf, suhail2022light, wang2022progressively}, leverage LFNs to produce comparable results to that of NeRFs, with tradeoffs between visual fidelity and speed.

Back propagation of gradients through a neural field MLP provides an ability for networks to refine parameters such as pose. While bundle-adjusting radiance fields~\cite{lin2021barf} refines poses during the formation of a radiance field, using a 
NeRF as a supervisory signal for absolute pose regression~\cite{chen2021direct} shows advantages in accuracy around convex and extended scenes. NeRFs do not perform well on few-shot datasets and require dense image coverage of a scene to create high fidelity results, however once generated it is possible to perform accurate pose regression on a minimal dataset~\cite{moreau2022lens}. Of key note is that the refinement or determination of pose can be performed in addition to estimation of other parameters within the neural field, such as shape, reflectance functions or illumination~\cite{boss2022-samurai}.

Joint learning of camera intrinsics and neural fields show improved extrinsics estimation. Wang et al.~\cite{wang2021nerf--} demonstrate an ability to jointly learn focal length and extrinsics, achieving similar results to traditional methods like COLMAP~\cite{schoenberger2016sfm}.

Jeong et al.~\cite{jeong2021selfcalib} jointly learn a complex non-linear distortion camera model with the neural field  in several stages, allowing the framework to learn a simple pinhole model follwed by complex components including non-linear distortion parameters. In essence, this curriculum learning approach enables the network to obtain correct scene geometry without warping the scene to agree with camera distortion.
\begin{figure*}[t]
	\centering  
	\includegraphics[width=\textwidth]{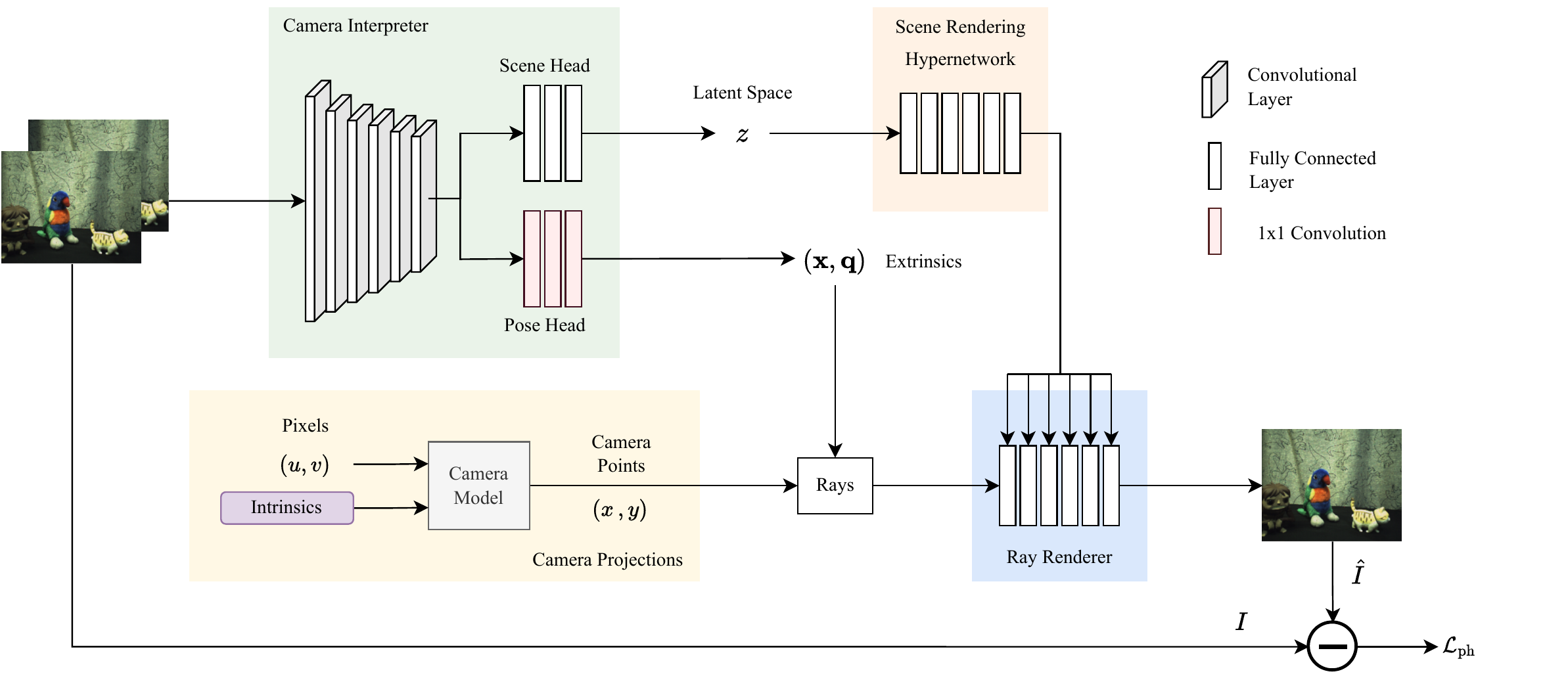}
	\caption{The proposed NOCaL network architecture is made up of 4 separate sub-modules. The camera interpreter (green) encodes input frames into a scene-description latent space \(\mathbf{z}\). The scene rendering hypernetwork (orange) uses the latent space to produce weights for the light field network (blue) which renders the scene captured by the input images. The camera model (yellow) has learnable parameters to be able to estimate the camera intrinsics used in capturing the input frames, and generates rays used by the light field network.}
	\label{fig_network_arch}
\end{figure*}

Hypernetworks \cite{ha2016hypernetworks} allow one network to produce weights for other networks that can perform additional tasks. The framework presented by von Oswald et al.~\cite{von2019continualhyper} has the ability to retain a vast amount of memory for multitask learning using a hypernetwork.  This benefits the individual networks and allows for reuse, owing to the commonalities between the learning tasks.

Sitzmann et al.~\cite{sitzmann2021lfns} also employs hypernetworks to leverage the ability to learn latent-based embeddings to produce a rendering network that can render the specific scene represented by latent embedding. Unlike prior works which focus on single scene, this network has the ability to render multiple scenes based on a single image.


%% file: sections/03_methods.tex

\section{Method}
\label{sec:methods}

\subsection{Network Architecture}
The two main learnable parts of the proposed network are the encoding network and the hypernetwork. These two parts work together to be able to jointly learn pose and scene geometry from input images. Details for implementation are discussed in Section~\ref{sec:implement}.

\subsubsection{Encoding Network}
The camera interpreter portion of our pipeline serves as an encoding network, which converts a pair of input frames from a specific camera to a pose and a latent space which represents the scene. This encoding network is built using a CNN structure with two separate heads at the end. The pose head outputs 3 values corresponding to translation and 6 values representing a continuous rotation space, of the form proposed by~\cite{zhou2019continuity}. 

The scene head outputs a 256-dimensional latent space \(\mathbf{z}_i\) which contains the information required to be able to build a rendering of the scene that the two input frames comes from. 

\subsubsection{Hypernetwork}
The hypernetwork $\mathbf{\Psi}$ uses the latent space $\mathbf{z}_i$ to output the weights for the rendering network $\boldsymbol{\psi}_i$,
\begin{equation}
    \mathbf{\Psi}(\mathbf{z}_i) = \boldsymbol{\psi}_i.
\end{equation}
Here the encoder has already distilled scene-specific information from the images, invariant to the camera. The operation of the rendering network is on the level of light rays, requiring no camera model to generate new images. In this way, the hypernetwork is able to be used as a tool for multiple cameras, giving the rendering network an initialisation which may be used to train the extrinsics and intrinsics of unknown vision sensors.

\subsubsection{Neural Fields for Supervision}
We utilise the rendering from the light field network to train pose and camera intrinsics. There are a few key benefits with using view synthesis from a neural light field. Firstly, it is ray based, providing a general model for all cameras, and resulting in novel views of sufficient visual fidelity ideal for use as supervision of odometry. Secondly, the implementation is fully differentiable which allows for an end to end system to be developed in which the input image into the neural field can be learnt. We utilise this second notion to learn camera parameters through a differentiable camera model.

The chosen light field network approximates a continuous scene in the form of an MLP $\mathbf{F}_{\boldsymbol{\psi}}:(\mathbf{o}, \mathbf{d}) \mapsto \mathbf{c}$ with weights $\boldsymbol{\psi}$. This formulation uniquely maps the ray direction $\mathbf{d}$ through some origin $\mathbf{o}$ using a Pl\"ucker coordinate encoding to a colour $\mathbf{c}$. As noted by Sitzmann et al.~\cite{sitzmann2021lfns}, whilst providing a compact and unique encoding, complex phenomena such as occlusion are not readily dealt with. The utilisation of a preceding frame to inform the hypernetwork and sequential camera motion enables the rendering network to avoid this shortcoming by evolving the scene representation over a trajectory.

\subsection{Camera Modelling}
Camera parameters are estimated in two parts: intrinsics, consisting of focal lengths $(f_x, f_y)$ and principal points $(c_x, c_y)$, and a distortion model. Together these describe a large family of cameras, except those not well described by a pinhole projection.
\subsubsection{Focal Length Estimation}
The estimation of the focal length is performed through back-propagation of the rendering MLP. Setting the focal length as a tuneable parameters that can be optimised allows the network to change the physical model of the camera as it generates scene geometry. The convergence of the focal length to correct values involves careful choice of initial value and the learning rate of the focal length in comparison to the rest of the network. The focal length is initialised as the image width in pixels, which will typically place the focal length within a suitable error margin such that the network will not get stuck in a local minima.

Given focal length is directly correlated to the scale of the geometry seen on the sensor, and we seek to jointly learn a representation of the scene and the camera values, small changes to focal length compared to geometry can trap the network in local minima. To this end, we use a higher learning rate to converge camera parameters prior to the network learning substantial scene geometry. 

\subsubsection{Implicit Non-Linear Distortion}
\begin{figure}
	\centering  
	\includegraphics[width=\columnwidth]{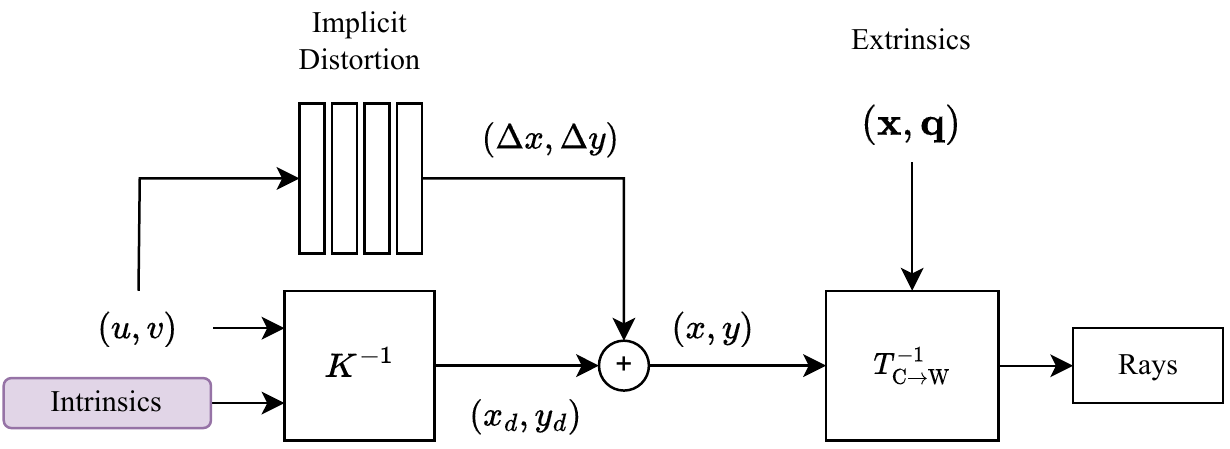}
	\caption{Model of camera parameter estimation. The pixel coordinates $(u,v)$ and the intrinsics $K$ are first used to calculate the distorted points on the image plane $(x_d,y_d)$. A distortion network takes as input $(u,v)$ and determines the distortion values on a per pixel basis and produces a \(\Delta x\) and \(\Delta y\). These corrections are applied to the points on the image plane producing undistorted points $(x,y)$, which are converted to rays in world space by the homogenous transformation $T_{C\to W}^{-1}$.}
	\label{fig_method_camera}
\end{figure}

To deal with generality of cameras, and to avoid the limitations of any single camera model, we model the non-linear distortion using an MLP, $\mathbf{D}(u,v) = (\Delta x, \Delta y)$. The MLP determines a \emph{correction} to coordinates on the camera plane using pixel coordinates $(u,v)$. This pixel coordinate-based MLP effectively is modelling a differentiable, smooth and continuous distortion function of sufficient complexity to encompass most cameras covered by a pinhole model. Figure~\ref{fig_method_camera} shows a diagrammatic representation of the MLP based distortion function in conjunction with the camera model.

\subsection{Semi-Supervised Learning}
In previous unsupervised learning of odometry work the camera parameters were required to be known and images undistorted prior to entering the pipeline~\cite{zhou2017unsupervised}. In this work we proposed learning the camera parameters in addition to relative pose within the pipeline. This adds a substantial degree of complexity to the network, which is less constrained. 

To reintroduce some constraints, a small amount of labelled data is used to semi-supervise the network. This allows us to directly impose a loss on the encoder, instead of having to backpropagate through the hypernetwork and allows us to confine the pose to metric terms. Previous unsupervised works had to scale the results after training to recover a metric scale~\cite{zhou2017unsupervised}. By imposing the learning of the camera parameters, we enable direct recovery of relative pose by avoiding scale ambiguity.

The trade-off between needing to know the camera parameters and needing a small amount of labelled data is often a preferable one. Using a platform such as a robotic arm enables a set of ground truth poses to be acquired irrespective of camera installed on-board along a pre-defined trajectory. Novel cameras may require extensive processes to acquire accurate data for direct calibration, which may be costly to obtain in large volumes. Using a small amount of this data, it enables the network to extend automatically and generalise to new calibrations. 

We demonstrate that this semi-supervised approach outperforms both a fully supervised and unsupervised approach. See Table. \ref{tab:odom} for results.

\subsection{Training Losses}
Similar to other works in neural rendering, we employ a photometric loss term $\mathcal{L}_\mathrm{ph}$ as the primary loss function of our network between ground truth $\mathbf{c}$ and predicted $\mathbf{\hat{c}}$ pixel colours. This is calculated for all rays $r\in\mathcal{R}$, where $\mathcal{R}$ is the set of rays captured by an image,
\begin{equation}
    \mathcal{L}_\mathrm{ph} = \sum_{r\in\mathcal{R}} |\mathbf{c}-\mathbf{\hat{c}}|_2^2.
\end{equation}

Where images have labelled poses during training, denoted as $\mathcal{I}'$, we enforce simple $L_2$-norms between the translations $\mathbf{x}\in\mathbb{R}^3$ and rotation matrices $\mathbf{R}\in\mathrm{SO}(3)$,
\begin{equation}
    \mathcal{L}_\mathrm{trans} = \sum_\mathcal{I'}|\mathbf{x}-\mathbf{\hat{x}}|_2^2,
\end{equation}
\begin{equation}
    \mathcal{L}_\mathrm{rot} = \sum_\mathcal{I'}|\mathbf{R}-\mathbf{\hat{R}}|_2^2.
\end{equation}
Finally, we encourage the latent space to have a mean of zero by assuming a Gaussian prior~\cite{sitzmann2021lfns},
\begin{equation}
    \mathcal{L}_\mathrm{enc} = \sum_\mathcal{I}\mathrm{mean}(\mathbf{z}).
\end{equation}
The overall loss function hence encompasses a loss from rendering, any available pose supervision and an imposed constraint to the latent space
\begin{equation}
    \mathcal{L} = \lambda_\mathrm{ph}\mathcal{L}_\mathrm{ph} + \lambda_\mathrm{trans}\mathcal{L}_\mathrm{trans} + \lambda_\mathrm{rot}\mathcal{L}_\mathrm{rot} + \lambda_\mathrm{enc}\mathcal{L}_\mathrm{enc}.
    \label{eqn:losses}
\end{equation}

\subsection{Curriculum Learning}
Given the challenges of jointly estimating scene and camera parameters, we employ a curriculum learning approach to sequentially recover camera parameters within an evolving neural scene. Initially, the encoding and hypernetwork are trained with fixed initial camera intrinsics, providing a rough low-frequency representation of the scene. This rough representation provides sufficient geometry to begin supervising the camera model. Prior to the geometry being fully converged, we enable the tuning of focal length in a simple pinhole model, allowing for adjustment of scene scale. Finally, the full implicit distortion model is added, giving a metric and geometrically representative scene representation, and providing a camera calibration. Learning in this way lets the network avoid local minima as higher frequency scene information is learnt.

%% file: sections/04_results.tex
\section{Results}
\label{sec:results}

\subsection{Datasets}
We demonstrate the results for the proposed system on both real and synthetic data, showing the system working for multiple cameras and scenes. 
The real world dataset used was part of the LearnLFOdo Dataset~\cite{digumarti2021unsupervised}. While this dataset is captured with a light field camera and this work is focused on monocular imaging, the dataset is the same as monocular when only the centre image is taken. This dataset contains 45 separate scenes and camera trajectories using the same camera captured by a UR5e robotic arm, providing accurate ground truth data to test the generalisation capabilities of NOCaL to new scenes.

Simulated results were also utilised to provide comparison between multiple cameras with defined intrinsics and distortion. This allowed for repeatable scene configurations and trajectories with multiple cameras.

\subsection{Implementation Details}~\label{sec:implement}
Both the hypernetwork and LFN are 6-layer MLPs with ReLU activations. The hypernetwork has 256 units per layer to allow for enhanced generalisation, while the LFN has 128 units. To encode the input image pairs a 7-layer CNN with kernel size of 3 is selected to maintain resolution of features. The pose head is a 3-layer CNN of 1x1 convolutions, while the scene head is a 3-layer fully connected network of 256 unit wide layers. Finally, the distortion model is an 8-unit wide MLP with 4-layers. We select the following hyperparameters, as laid out in Eqn.~\ref{eqn:losses}, $\lambda_\mathrm{ph} = 100$, $\lambda_\mathrm{trans} = 30$, $\lambda_\mathrm{rot} = 20$, $\lambda_\mathrm{enc} = 1\times10^{-6}$.

Four separate learning rates were used to train the separate parts of the model, one for each of the separate sub-modules; the intrinsics, the distortion, the hyper rendering network, and the encoding network. As the network jointly optimises the separate sections if one trains faster than another, it can lead to modules getting stuck in a local minima. Weighting the separate parts through the learning rates is critical to ensure the framework as a whole converges appropriately. The chosen learning rates were, $5\times10^{-5}$, $8\times10^{-5}$, $5\times10^{-6}$, $5\times10^{-1}$ for the encoder, hypernetwork, distortion model and intrinsics respectively. Adam optimisers were used for all separate learning rates. All layers were initialised with an Xavier uniform distribution~\cite{glorot2010understanding}. 

\begin{figure}
	\centering  
	\includegraphics[width=\columnwidth]{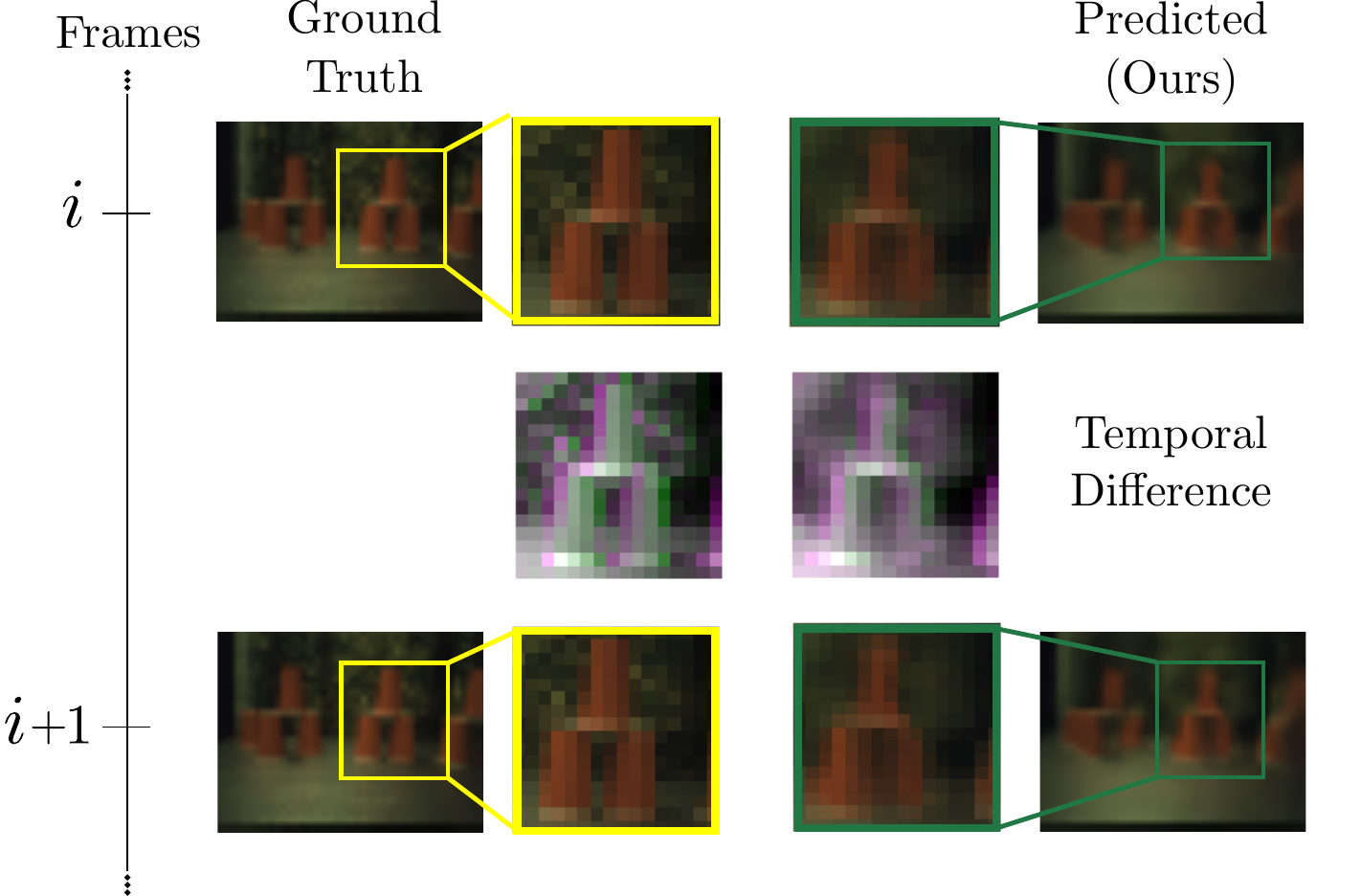}
	\caption{View synthesis from the learnt LFN on a test scene. A pair of input images at times $i$, $i+1$ are shown on the left, with the corresponding predicted views on the right closely matching in visual appearance. The temporal difference frames shows the motion through time, it can be seen that the predicted and measured motion are very similar. Animations of this motion will be made available on the project website.}
	\label{fig_supervision_qual}
\end{figure}

\begin{table*}
\centering
\caption{Evaluating camera parameter estimation}
\begin{tabular}{ |c|c|c|c|c|c|c| } 
 \hline
 \rule{0pt}{9.5pt}\multirow{2}{*}{Camera} & \multicolumn{2}{|c|}{Initial Error} & \multicolumn{2}{|c|}{NOCaL (ours)} & \multicolumn{2}{|c|}{COLMAP~\cite{schoenberger2016sfm}}\\ 
 \cline{2-7}\rule{0pt}{9.5pt}
 & $f$ [px] & Error $\overline{\Delta r}$ & $f$ [px] & Error $\overline{\Delta r}$ & $f$ [px] & Error $\overline{\Delta r}$\\
 \hline
 Focal: 600px, no distortion & 640.0 & 0.0225 & 601.1 & 0.0008 & 599.4 & 0.0003\\
 Focal: 600px, large distortion & 640.0 & 0.0440 & 595.3 & 0.0186 & 602.1 & 0.0010\\ 
 \hline
\end{tabular}
\label{tab:cam_params}
\end{table*}

\begin{table*}[t]
\centering
\caption{Evaluating odometry performance on captured and rendered imagery}
\begin{tabular}{ |c|c|c|c|c|c|c|c|c| } 
 \hline
 \rule{0pt}{9.5pt}\multirow{2}{*}{Method} & \multirow{2}{*}{Labelled Images} & \multirow{2}{*}{Unlabelled Images} & \multicolumn{3}{|c|}{Translation Error [m]} & \multicolumn{3}{|c|}{Rotation Error [degrees]}\\
 \cline{4-9}
 \rule{0pt}{9.5pt}&&& Mean & STD & RMSE & Mean & STD & RMSE \\
 \hline
 \multicolumn{9}{|c|}{Odometry accuracy on captured indoor imagery}\\
 \hline
 Fully supervised & 800 & 0 & 0.025 & 0.009 & 0.027 & 1.553 & 1.847 & 2.414\\ 
 Unlabelled calibrated~\cite{zhou2017unsupervised} & 0 & 8000 & 0.029 & 0.016 & 0.033 & 1.522 & 0.969 & 1.808\\
 NOCaL (ours) & 800 & 7200 & \textbf{0.020} & \textbf{0.008} & \textbf{0.022} & \textbf{0.412} & \textbf{0.295} & \textbf{0.505}\\
 \hline
 \hline
 \multicolumn{9}{|c|}{Ablation study using rendered indoor imagery with camera distortion}\\
 \hline
 Ours no intrinsics or distortion & 100 & 900 & 0.157 & 0.060 & 0.168 & 8.026 & 9.180 & 12.194\\ 
 Ours no distortion & 100 & 900 & 0.147 & \textbf{0.053} & 0.156 & 4.790 & 2.209 & 5.275\\ 
 Ours full & 100 & 900 & \textbf{0.145} & 0.054 & \textbf{0.154} & \textbf{4.024} & \textbf{1.971} & \textbf{4.481}\\ 
 \hline
\end{tabular}
\label{tab:odom}
\end{table*}

\subsection{Scene Reconstruction}
As shown in Fig.~\ref{fig_supervision_qual}, the framework is able to produce novel views of scenes it has not been trained on. Given a pair of inputs with some motion between frames, a pair of predicted frames can be retrieved from the network. As these views are used as the supervisory signal for the rest of the network, the temporal difference or motion between the predicted frames should reflect the same motion between the input frames. This is the case in Fig.~\ref{fig_supervision_qual}, which indicates that the network can be supervised with this signal. The reconstructions have lost some of the high frequency scene content, however the reconstruction quality is not critical, but rather how well the renderings can supervise the motion between the frames.

\subsection{Camera Modeling}
NOCaL is able to recover accurate camera intrinsics, close to ground truth and those attained by traditional methods. Table~\ref{tab:cam_params} demonstrates an ability to recover comparable results to COLMAP~\cite{schoenberger2016sfm} in the absence of distortion, validating the case of an ideal pinhole camera by a significant reduction in error. We compare results based on the mean radial shift per pixel $\Delta r = \sqrt{\Delta x^2 + \Delta y^2}$. We sample the radial distortion function used by COLMAP on a grid to obtain a comparable result owing to the formulation of the distortion network in this work.

We note that while our method is able to significantly reduce the error in the case of large distortion, the fixed camera model used by COLMAP provides a better approximation to the continuous radial distortion profile. As we are also jointly learning focal length, some error in focal length is taken up within the distortion network leading to additional error. Additional constraints could be imposed on the implicit distortion model in the future, though we note this did not affect achieved odometry performance.

\subsection{Odometry Results}
Odometry results for NOCaL are shown in Table~\ref{tab:odom}. We compared NOCaL to two other odometry methods: a fully supervised approach with labelled imagery, and an unsupervised approach based on \cite{zhou2017unsupervised} that requires the camera to be calibrated and imagery rectified. 
The unsupervised method was provided with similar numbers of unlabelled images, around 8000, representative of the availability of unlabelled imagery in practical scenarios. 

While the unsupervised approach did not require any labelled data, it is scale ambiguous, with the results needing to be correctly scaled before an error can be calculated. NOCaL does not require such scaling as the labelled data during training establish scale based on semantic information in the image. Furthermore our framework does not require camera calibration or rectification.

Perhaps surprisingly, our zero-calibration approach outperformed both fully supervised and calibrated unsupervised methods. This is partially explained by the amount of training data available to each method: NOCaL and the fully supervised approach were provided with 800 labelled images, but NOCaL also has the benefit of additional unlabelled imagery. In the case of the unsupervised calibrated method, we hypothesise that our freeform distortion model did a better job of describing the camera distortion compared to the parameteric model employed in typical camera calibration.

We performed an ablation study to measure the effectiveness of our camera model -- the results are shown at the bottom of Table~\ref{tab:odom}. We tested the full NOCaL, NOCaL without a distortion model, and a version that does not adjust the camera intrinsics or distortion model. This study employed rendered imagery simulating a camera with realistic and known radial distortion. Intrinsics were initialised close to but not exactly matching correct values, in line with a typical imaging scenario. The study shows both the distortion model and intrinsic refinement play an important role in NOCaLs strong odometry performance. The presented error metrics were calculated using~\cite{grupp2017evo}.

\subsection{Training and Inference Time}
Typical training time for NOCaL on the LFodo dataset~\cite{digumarti2021unsupervised} was approximately $1.5$ hours. Training time was tempered by use of the LFN for rendering and use of down-sampled training images. Compared to a ray marching alternative such as NeRF\cite{mildenhall2021nerf}, an LFN is much faster as it only needs to sample each ray once. 

As the rendering and the odometry can be uncoupled, inference is performed on the odometry network alone, yielding an inference time of $16.9$ ms, compatible with real-time applications. Performing inference of the full framework takes $33.9$ ms. The networks were trained and timed on an NVIDIA RTX 3060 12GB GPU.

%% file: sections/06_conclusions.tex
\section{Conclusions}
\label{sec:concl}

We presented NOCaL, a framework that jointly learns odometry, camera parameters and visual appearance in an end to end fashion. This framework generalises to a large range of cameras that can be modelled with a pinhole projective model and freeform distortion map. We demonstrated our method learning odometry on previously unseen and uncalibrated cameras, and an ablation study established the importance of learning camera parameters for this task. Our approach requires only a small amount of labelled data, allowing it to provide metric results with scale, and benefits from the availability of large amounts of unlabelled data from both a newly introduced camera and the vast quantities of existing unlabelled data from existing cameras. 


Future work will entail extending the camera model to work for more general camera geometries, including fish-eye lenses and multi-aperture cameras. Deployment on robotic platforms undergoing extended operation with online updates to the camera model represents a logical next step in validating the approach. Finally, we aim to extend the work to support time of flight, event, and more general computational imaging devices, working towards truly autonomous integration of emerging camera technologies.